# Exploring the Relationship Between Team Institutional Composition and Novelty in Academic Papers Based on Fine-Grained Knowledge Entities

Ziling Chen, Chengzhi Zhang*, Heng Zhang, Yi Zhao, Chen Yang, and Yang Yang
Department of Information Management, Nanjing University of Science and Technology, Nanjing, China

## Abstract

**Purpose** – The composition of author teams is a significant factor affecting the novelty of academic paper. Existing research lacks studies focusing on institutional types and measures of novelty remained at a general level, making it difficult to analyze the types of novelty in papers and provide a detailed explanation of novelty. This study takes the field of Natural Language Processing (NLP) as an example to analyze the relationship between team institutional composition and the fine-grained novelty of academic papers.

**Design/methodology/approach** – Firstly, author teams are categorized into three types: academic institutions, industrial institutions, and mixed academic and industrial institutions. Next, the authors extract four types of entities from the full paper: methods, data sets, tools and metric. The novelty of papers is evaluated using entity combination measurement methods. Additionally, pairwise combinations of different types of fine-grained entities are analyzed to assess their contributions to novel papers.

**Findings** – The results of the study found that in the field of NLP, for industrial institutions, collaboration with academic institutions has a higher probability of producing novel papers. From the contribution rate of different types of fine-grained knowledge entities, the mixed academic and industrial institutions pay more attention to the novelty of the combination of method indicators, and the industrial institutions pay more attention to the novelty of the combination of method tools.

**Originality/value** –This paper explores the relationship between the team institutional composition and the novelty of academic papers, and reveals the importance of cooperation between industry and academia through fine-grained novelty measurement, which provides key guidance for improving the quality of papers and promoting industry-university-research cooperation.

* Corresponding author.
E-mail address: zhangcz@njust.edu.cn (C. Zhang).

## 1 Introduction

Innovation is the basis for promoting scientific development (Shibayama et al.,2021), and it is of great significance to conduct scientific and effective measurement and evaluation on it. The concept of Innovation originated from the Innovation Theory proposed by Schumpeter. Schumpeter(1934) believed that innovation is a revolutionary change that must be able to create new value. At present, the academic community has not reached a consensus on the definition and connotation of academic literature innovation, but the discussion on it all reflects the essential feature of "new", which also makes the current research focus on novelty. The importance of novelty in academic papers cannot be ignored. Novelty is essential in academic papers, reflecting their standard and serving as a crucial indicator for evaluation. According to Mishra et al. (2016) , the novelty of academic papers manifests in the introduction of new perspectives, encompassing both previously unheard-of viewpoints and the augmentation of existing established viewpoints. Sarah et al. (2014), approaching from the perspective of knowledge combinations, posit that novelty arises from the development of atypical combinations of previously existing knowledge.

The composition of a scholar team has a significant impact on the novelty of academic papers. The composition of a scholar team encompasses factors such as team size, authors' gender, and collaboration types, which to some extent influence the level of innovation in papers. For instance, Yang et al. (2022) found that the higher the gender balance in a team, the more prominent the advantage in novelty. Lee et al. (2015) discovered that there is an inverted U-shaped relationship between team size and the novelty of academic papers. Studying the relationship between team institutional composition and novelty can promote the improvement of team effectiveness. Previous analyses of the novelty of academic papers have typically focused on factors such as team size, gender composition ratio, and the presence of international collaboration. However, there has been a lack of analysis from the perspective of the institutions to which author teams belong. This oversight neglects the impact of institutions on the novelty of research outcomes, leading to an imbalance in the current research perspectives on the relationship between scholar team composition and paper novelty. Historically, universities have been central to the national innovation system, renowned for generating cutting-edge scientific knowledge and advancing technology (Nelson, 1993; Kang & Motohashi, 2020). As technology progresses and knowledge spreads, industrial institutions are now investing heavily in research and development, collaborating with academia to drive the application of new technologies (Chai & Shih, 2016). Simultaneously, academia employed various means to disseminate newly produced knowledge to industry, thus driving economic growth (Mowery et al., 2015). Previous studies suggested that industrial knowledge primarily originated from academia (Bikard and Marx, 2020), prompting industries to prioritize indigenous innovation efforts (Karhade and Dong, 2021). Presently, industry is evolving from being consumers of external knowledge for technology development to becoming significant creators of knowledge (Suzuki et al., 2017; Jong & Slavova, 2014). Furthermore, as industrial institutions play an increasingly prominent role in publishing academic papers, disparities between industrial and academic institutions in research direction, resources, and methodologies become more apparent. Academic research has always strived for novelty, and due to institutional differences, various types of institutions may exhibit their own bias towards novelty. For example, academic institutions may emphasize theoretical innovation, while industrial entities may prioritize applied innovation (Mansfield, 1998; Mansfield, 1991).

Analyzing the potential significant differences in the inclination towards novelty types among different types of institutions can enhance our understanding of the characteristics and strengths unique to each type of institution. Such analysis provides valuable insights and directions for interdisciplinary collaboration and communication.Current research primarily focuses on the macro-level assessment of the overall innovativeness of scientific literature, while overlooking the innovation level of specific knowledge elements (Wang et al., 2021). The previous methods of measuring novelty did not involve fine-grained features such as type bias, making it difficult to analyze type bias and unable to further explain the measurement results in detail. Previous methods of measuring novelty have not addressed nuances like bias towards specific types of novelty, which hinders detailed analysis of novelty types in papers and nuanced interpretations of novelty measurement outcomes.Traditional novelty measurement methods, relying on entity combinations, quantitatively assess novelty by determining the proportion of atypical entity combinations in academic papers. Expanding on this, we can extract various types of fine-grained knowledge entities and analyze atypical entity combinations. This helps calculate the proportion of different types of fine-grained knowledge entities within these combinations. By studying the contributions of different types of fine-grained entity combinations to paper novelty from a fine-grained entity perspective across different institutions, disparities in fine-grained novelty bias among institution types can be revealed.

This research makes two main contributions:

Firstly, this study focuses on the composition of different institutions in industry and academia, and deeply analyzes the differences between these different types of team institutional compositions on the novelty of academic papers. By delving into the relationship between team institutional composition and paper novelty, this study provides a fresh perspective for related research.

Secondly, this study conducts analysis of the measurement of novelty in academic papers. It introduces a novelty bias measurement method based on fine-grained knowledge entities within scientific entities. By analyzing the contributions of different fine-grained entity combinations, the measurement of novelty has evolved from the traditional simple binary and ordinal levels to different types of novelty bias levels.

## 2 Related work

The aim of this paper is to explore the relationship between team institutional composition and academic paper novelty. This section provides a review of existing research from two perspectives: the novelty measurement of academic paper and the association between scholar team composition and academic paper novelty.

### 2.1 Novelty Measurement of Academic Paper

Schumpeter (1934) proposed in his book "The Theory of Economic Development" that innovation is the introduction of a new production function, bringing new combinations of production factors and conditions into the production system that have not been seen before. This theory, universally recognized since its inception, has spurred a series of exploratory studies deepening our understanding of innovation's sources and characteristics. Many researchers (Foster et al., 2015; Mukherjee et al., 2016) believed that novelty primarily arised from novel combinations of previously uncombined elements or established elements with new concepts. This process builded upon existing knowledge, creating new knowledge while remaining within mainstream research fields and paradigms. . Yan et al. (2020) suggested that the novelty of academic papers, from the perspective of combining scientific processes, stems from combining knowledge elements in a new way. Following the logic of combinatorial innovation, scholars no longer view novelty as "unmeasurable," but instead have devised metrics to gauge the innovativeness of academic papers or the extent to which they innovate (Kaplan & Vakili, 2015). Novelty measurement is a challenging and cutting-edge direction in informetrics (Wang et al., 2023). They conceptualized the novelty of academic papers as the degree to which different fine-grained knowledge entity in academic papers are connected in new or unusual combinations. This approach has been widely used in various studies on novelty measurement, where researchers primarily analyzed citation relationships between externally related literature or internal textual semantic information such as topics, innovative sentences, keywords, and knowledge entities.

In research on measuring novelty based on external relationships between literature, a significant study involves the combination of journal references. Uzzi et al. (2013) proposed a method that captures the combination process of academic papers by calculating the relative commonality of journal pairs cited in papers. This strategy has been applied and adapted in a series of subsequent related works due to its comprehensiveness and originality. Wang et al. (2017) viewed scientific research as a combinatorial process where novelty is the exploration of combining new knowledge with existing knowledge. They measured the novelty of science based on whether papers combine reference journals for the first time. Wang et al. (2024) utilized citation relationships between journals from different disciplines to construct a citation matrix. They then transformed it into a similarity matrix using cosine normalization, thus establishing a reliable indicator for identifying interdisciplinary breakthrough innovations.

In addition, for evaluating novelty based on the intrinsic semantic knowledge of papers, in the early years, it was mainly done through expert evaluation and peer review methods. However, with the rapid increase in the number of academic papers, this method cannot effectively handle big data. Therefore, many scholars in different fields use knowledge elements such as topics, innovative sentences, keywords, entities, and other information from text to measure novelty. Boudreau et al. (2016) proposed an individual novelty measurement for research funding proposals based on new combinations of Medical Subject Headings (MeSH) terms. Mishra et al. (2016)

explored the relationship between MeSH terms and the novelty of papers in biomedical research. They suggested that individual MeSH terms in biomedicine may not show novelty well, but pairs of MeSH terms can better reflect the novelty of papers. The most influential papers often introduce some novel combinations (atypical combinations) on the basis of traditional combinations (typical combinations). Foster et al. (2015) extracted chemical entities annotated in official biomedical annotations from biomedical papers, constructed a chemical knowledge network using entity combinations, and defined combinations of knowledge entities in different clusters of the knowledge network as innovative. Liu et al. (2022) used the BioBERT model to extract biological entities from medical papers in the PubMed database and proposed a method to calculate the novelty of each academic paper based on the proportion of rare entity combinations. Luo et al. (2022) extracted research methods and research questions from abstracts of academic papers. Using the perspective of combinatorial innovation and considering time factors and semantic distances, they developed a new indicator for measuring the novelty of academic papers.

In summary, existing research on the novelty measurement of academic papers mainly revolved around two approaches: novelty measurement based on reference combinations and novelty measurement based on textual content. Studies in this area primarily focused on measuring the novelty of combinations such as journal references, citation pairs, topic terms, and entity pairs, providing valuable insights for this research.

## 2.2 Relationship Between Scholar Team Composition and Academic Paper Novelty

In the past, scholars have explored the relationship between team composition and the novelty of academic papers from perspectives such as team size, gender composition, leadership relationships, and international collaboration. For instance, Lee et al. (2015) investigated the relationship between team size and paper novelty, revealing a inverted U-shaped relationship, moderated by the diversity of research fields and task diversity. Yang et al. (2022) examined the impact of gender diversity on paper novelty by considering each paper as a research team. They found that papers produced by gender-diverse teams were more novel and academically influential compared to papers from teams of the same gender with equivalent team size. Xu et al. (2022) explored the relationship between team leadership structure and paper novelty. They categorized team members into leaders and supporters based on their contributions to the research and found that a flatter team structure was more conducive to producing novel outcomes. The question of whether international collaboration leads to more novel outcomes has also been a focal point of interest. Wagner et al. (2019) analyzed data from Web of Science and Scopus in 2005 and concluded that international collaboration does not necessarily promote novel outcomes; instead, transaction costs and communication barriers associated with international collaboration may inhibit the production of novel outcomes. Liu et al. (2021) focused on the impact of international collaboration and first-time collaboration on team performance during the COVID-19 pandemic. They found that during the pandemic, teams with more first-time collaborations produced more novel academic papers, while international collaboration did not hinder the production of novel outcomes as it did during normal times.

From the above review, it can be observed that existing research has mostly focused on individual attributes within teams, such as team size, gender composition, leadership relationships, and the presence of international collaboration, when exploring the relationship with team outcome novelty. Currently, few scholars have categorized teams based on the institutional affiliations of individual authors and investigated the relationship between different types of institutional collaboration and team outcome novelty. This paper aims to address this gap by examining team composition from the perspective of authors' institutional affiliations, categorizing them into industrial and academic sectors. It further analyzes the relationship between different types of institutional collaboration and team outcome novelty.

## 3 Methodology

This study aims to analyze the relationship between team institutional composition and academic paper novelty. We utilized an automated scientific entity recognition method to extract technology-related entities from NLP papers. Subsequently, we measured the novelty of academic papers based on combinational innovation theory using knowledge entities as units of analysis. The framework of our study is illustrated in Figure 1. We divided the study into four steps. The first step involved dataset construction, where we gathered papers in the NLP field, extracted basic information such as publication year, author names, and affiliations, and categorized team institutional composition based on authors' affiliations. In the second step, we used the SciBERT+BiLSTM (cascade) model to extract fine-grained entities from the full text of papers. The third step focused on calculating novelty scores for academic papers based on entity combinations to measure their novelty, examining the contribution of different entity types to novelty. The fourth step analyzed the relationship between team institutional composition and academic paper novelty, exploring how different types of fine-grained entity combinations contribute to the novelty of papers with varying team institutional compositions.

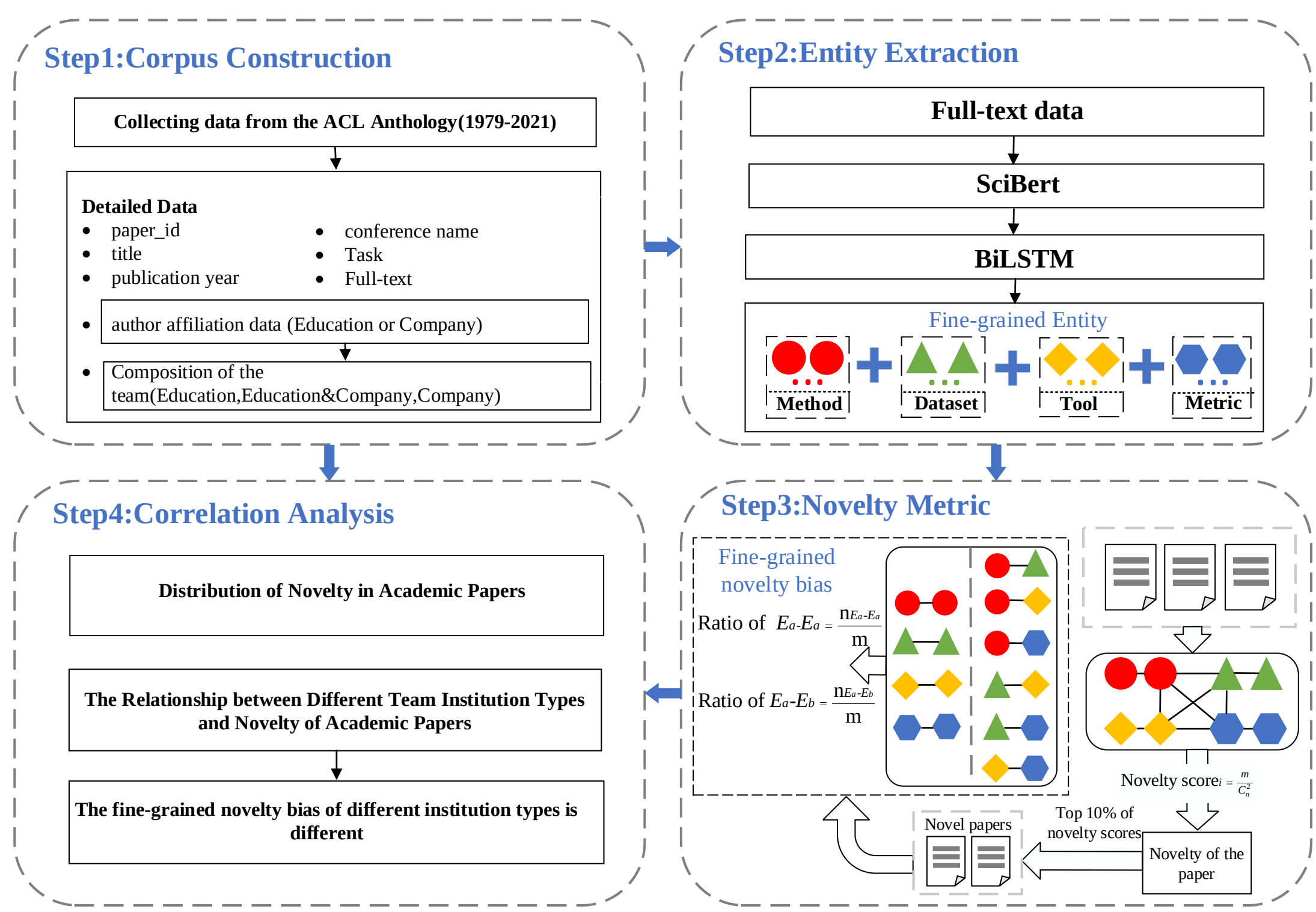

**Fig. 1. Framework of this study**

## 3.1 Data sources and processing

This study explores the relationship between institutional types in the NLP field and the novelty of academic papers. Research indicates that the computer science field values conferences as a publication venue more than any other academic field (Vrettas & Sanderson, 2015). The ACL Anthology[1] is a collection of academic papers in the fields of Computational Linguistics and Natural Language Processing, containing papers published at various important conferences and journals since the 1960s, providing a rich resource for NLP research (Bollmann & Elliott, 2020). Data for this study were collected by crawling three top-tier NLP conferences from the ACL Anthology: ACL (Association for Computational Linguistics), EMNLP (Empirical Methods in Natural Language Processing), and NAACL (North American Chapter of the Association for Computational Linguistics). These three major conferences are authoritative and representative in the NLP field, covering a wide range of research directions and topics in NLP. Zhang et al. (2024).analyzed the development trends of new technologies in the field of natural language processing based on datasets from these three major conferences.

PDF files of main conference papers from the three major conferences in NLP from 1979 to 2021 were collected from the ACL Anthology website. The pdf2htmlEX [2] tool was used to parse the PDF files into HTML files, and text data was extracted using rule-based matching to obtain full-text data from 15,337 papers. Metadata such as paper titles, authors, and publication years were also crawled from the website.

[1] https://aclanthology.org/
[2] https://pdf2htmlex.github.io/pdf2htmlEX/

In the field of NLP research, studies are typically task-oriented. Therefore, this paper references "The Oxford Handbook of Computational Linguistics" (Mitkov, 2022) and Wang & Zhang, 2020) to classify research tasks based on conference manual labels for controlling variables in subsequent related research. These classifications will be used to establish a regression model that controls variables in order to explore the relationship between organizational composition types and novelty. A total of 30 tasks were identified, including machine translation, semantics, information extraction, machine learning, syntactic analysis, sentiment analysis and opinion mining, resources and evaluation, natural language generation, dialogue, question answering, NLP in internet and social media, summarization, representation learning and text representation, text mining, NLP in multimodal and multimedia systems, information retrieval, discourse analysis, speech processing, NLP applications, syntax, grammar, word sense disambiguation, morphology, anaphora resolution, word segmentation, phonology, ethics and NLP, pragmatics, cognitive linguistics, and others. The "others" category refers to tasks in the NLP field that do not belong to the main task categories or have not been clearly classified. These tasks may have special requirements and challenges, necessitating further research to advance their development (Mitkov, 2022). The collected field content is summarized in Table 1.

**Table 1. List of paper fields collected by ACL Anthology**

| Paper | Example |
|---|---|
| Paper_id | 2020.acl-main.1 |
| Title | Learning to Understand Child-directed and Adult-directed Speech |
| Author | Gelderloos Lieke,Chrupaa Grzegorz ,Alishahi Afra |
| Author affiliations | tilburg university,tilburg university,tilburg university |
| Publication Year | 2020 |
| Conference Name | Acl |
| Task | Natural Language Generation |

### 3.2 Determination of Team Organizational Composition Types

This study examines the relationship between authors' institutional types and the novelty of academic papers. To begin, it is necessary to extract the institution names of each author and then categorize these institutions. A total of 15,337 papers from the years 1979-2021 were collected, resulting in 1,998 unique author affiliations. Following the approach of Manjunath et al. (2021), these affiliations were grouped into eight categories using the Global Research Identification Database (GRID) : government, education, corporations, institutions, healthcare, non-profit organizations, archives, and others. GRID collects institution names globally and supplements institution data such as country and city of location using Wikipedia[3] and GeoNames[4]. For institutions mentioned in the papers but not found in GRID, additional information was gathered online to supplement their data. Subsequently, a disambiguation process was conducted on the identified institutional data to determine unique names, standardizing capitalization, abbreviations, full names, and name order. The statistics of the institutions in the dataset are presented in Table 2.

[3] https://www.wikipedia.org/

[4] https://www.geonames.org/

Given that this study focuses on academic and industrial institutions, further classification was carried out based on the nature of the eight institution types. Academic and industrial institutions were categorized as academic and industrial institutions respectively. Notably, healthcare institutions like universities, medical schools, and research centers (e.g., Harvard Medical School, Mayo Clinic) were grouped under the academic category (Xu et al., 2022). Ultimately, academic and industrial institutions accounted for 96.5% of the total, minimizing data loss compared to the original dataset and ensuring experimental accuracy and reliability. Other institution types were excluded from the analysis, as the study specifically targets team organizational composition and excludes single-authored papers. A total of 14,812 academic papers were selected after screening, representing 1,901 unique affiliations: academic institutions dominated (67.5%, 1,285 institutions), while industrial institutions accounted for 32.5% (616 institutions). The institutional composition is summarized in Table 3. Subsequently, based on the institution type of each author, the team organizational composition categories of each academic paper were determined, including academic institutions (A&A), industrial institutions (I&I), and mixed academic and mixed academic and industrial institutions (A&I).

**Table 2. Distribution of Institutions Using the Manjunath et al.,(2021) Method**

| Category of Institution | Example | Count |
|---|---|---|
| Eduacation | Harvard University | 1277 |
| Company | IBM、Google | 616 |
| Institute | The Alan Turing Institute | 47 |
| Government | The Supreme People's Court | 18 |
| Non-profit institutions | Qatar Foundation | 15 |
| Archives | The European Library | 10 |
| Healthcare | Mayo Clinic | 8 |
| Independent | Independent Researcher | 7 |
| Total | | 1998 |

**Table 3. Distribution of Institutions Using the Xu et al.,(2022) Method**

| Category of Institution | Compose | Example | Count | Ratio |
|---|---|---|---|---|
| Academic institutions | Education, health care | Harvard University、Mayo Clinic | 1285 | 67.5% |
| Industry bodies | Firm | IBM、Google | 616 | 32.5% |
| Total | | | 1901 | 100% |

### 3.3 Novelty Measurement of Academic Papers Based on Fine-Grained Entities

This paper focuses on exploring the relationship between different team institutional compositions and the novelty of academic papers. The various institutional compositions have been defined earlier, and the following sections will primarily introduce the measurement of academic paper novelty. This study primarily focuses on measuring the novelty of academic papers based on fine-grained knowledge entity combinations, involving the extraction of fine-grained knowledge entities, the measurement of novelty in academic papers, and the quantification of fine-grained novelty bias in novel papers.

### 3.3.1 Extraction of Fine-Grained Knowledge Entities

#### (1) Data annotation

The measurement of novelty in academic papers in this study is based on the novelty measurement method of entity combinations. This method fundamentally conceptualizes the novelty of academic papers as the degree to which knowledge entities in academic papers are connected in new or atypical combinations, based on the theory of combinatorial innovation. Knowledge entities are considered the fundamental units and structural elements of knowledge in academic literature (Ding et al., 2013). In this research, accurately identifying knowledge entities is crucial for measuring the novelty of academic papers as it directly influences the evaluation of atypical knowledge combinations. In the field of NLP, effective novelty mainly manifests in methodological and technological innovations during the research process. Therefore, this study refers to the essential components of solutions corresponding to research problems in academic papers as method-related entities (hereafter referred to as "method entities").

The entity extraction methods and some data used in this study are derived from the research by Zhang et al. (2024). They focus on method-related entities and define four categories of entities: methods, datasets, metrics, and tools, with their meanings and examples provided in Table 4. They randomly selected 50 ACL papers to construct an entity recognition annotation dataset. Five researchers in the NLP field utilized the WebAnno[5] to complete the annotation task, annotating a total of 3,115 sentences containing method, dataset, tool, and metric entities.The dataset was split into training, validation, and testing sets in an 8:1:1 ratio.

**Table 4. Translation of Explanations and Examples for Four Entity Types**

| Type | Description | Example |
|---|---|---|
| Method | Algorithms, Models, etc. | SVM, LSTM, BERT, Adam, RNN |
| Dataset | Corpora, Lexicons, etc. | Brown Corpus, Penn Treebank, WordNet |
| Metric | Evaluation metrics | Accuracy, Precision, Recall, BLEU |
| Tool | Programming languages, Software, Open-source tools, etc. | Python, GIZA++, TensorFlow, PyTorch |

#### (2) Model Extraction

Based on the annotated dataset, they designed a cascaded model structure of SciBERT+BiLSTM for identifying method-related entities. Additionally, they

[5] https://webanno.github.io/webanno/

established baseline models comprising six variations:

- BERT: A pretrained language model based on the Transformer architecture, widely used for natural language processing tasks (Devlin et al., 2018).
- BERT+CRF: Combining the BERT pretrained language model with the conditional random field (CRF) sequence labeling model for named entity recognition tasks (Li et al., 2020).
- XLM-RoBERTa: A cross-lingual pretrained language model, leveraging a combination of RoBERTa and the cross-lingual model XLM, demonstrating excellent performance in multilingual natural language processing tasks (Conneau et al., 2019).
- XLM-RoBERTa+CRF: Combining the cross-lingual pretrained language model XLM-RoBERTa with the conditional random field (CRF) sequence labeling model for cross-lingual named entity recognition tasks (Choure et al., 2022).
- SciBERT: A pretrained language model tailored for scientific literature, pretrained on domain-specific corpora to better adapt to natural language processing tasks in the scientific domain (Beltagy et al., 2019).
- SciBERT+CRF: Combining the SciBERT pretrained language model with the conditional random field (CRF) sequence labeling model for named entity recognition tasks in scientific literature (Sahrawat et al., 2019).

According to Table 5, the results show that the SciBERT pre-training model performs better on the scientific entity recognition task than the general corpus pre-training model such as BERT and XLM-RoBERTa. The SciBERT+BiLSTM cascade model F1 scored the highest (86.27).Therefore, in this study, this model was applied to entity extraction tasks, extracting a total of 151,018 entities of methods, datasets, tools, and metrics from 14,812 full-text papers from the three major conferences.

**Table 5. Evaluation of models on our annotated dataset**

| Model | Precision | Recall | $F_1$ |
|---|---|---|---|
| BERT | 87.87 | 82.14 | 84.91 |
| BERT+CRF | 85.56 | 80.88 | 83.15 |
| XLM-RoBERTa | 84.34 | 84.87 | 84.61 |
| XLM-RoBERTa+CRF | 82.27 | 83.82 | 83.04 |
| SciBERT | 84.77 | 86.55 | 85.65 |
| SciBERT+CRF | 87.34 | 84.03 | 85.65 |
| SciBERT+BiLSTM(cascade) | 88.16 | 84.45 | **86.27** |

### 3.3.2 Novelty Measurement Method Based on Entity Combinations

To quantify the novelty of papers from the three major NLP conferences, it is essential to determine the semantic distance between each pair of method entities in terms of their respective entities. The Scibert+BiLSTM (cascade) model trained for entity extraction generates word embeddings for each entity, which are then used to calculate the cosine distance between the two semantic vectors corresponding to each entity in an entity combination, as per Formula (1). A greater cosine distance between the word embeddings of two entities indicates a lower semantic similarity, implying a lower degree of association and significant differences in meaning and semantic features between the two entities. By capturing the cosine distances for each pair of entities detected within 13,260,8161 entity combinations identified in papers from the three major NLP conferences, the study identifies the top 10% of entity combinations with the furthest distances as atypical combinations. The proportion of atypical combinations in academic papers serves as a specific metric for determining paper novelty. A higher Novelty score signifies a higher prevalence of atypical entity combinations in the paper, indicating high novelty, whereas a lower score suggests a lower proportion of atypical entity combinations, reflecting low novelty. Academic papers ranking in the top 10% based on novelty scores are defined as novel papers, while the rest are classified as conventional papers.

$$\text{Cosine distance}_{E_a\text{-}E_b} = 1 - \frac{E_a \cdot E_b}{\|E_a\| \cdot \|E_b\|} \tag{1}$$

$E_a$ and $E_b$ Representing the two entities in an entity pair; $E_a \cdot E_b$ is the dot product of $E_a$ and $E_b$, $\|E_a\| \cdot \|E_b\|$ represents the product of the Euclidean norms of $E_a$ and $E_b$.

$$\text{Novelty score}_{Doc_i} = \frac{m}{C_n^2} \tag{2}$$

In Formula (2), $Doc_i$ represents the academic paper; $n$ is the number of biomedical entities extracted from $Doc_i$; $C_n^2$ is the total number of combinations of two entities that can be extracted from the set of n biomedical entities extracted from $Doc_i$, i.e., the number of entity pairs generated by $n$ entities; $m$ denotes the count of entity pairs in $Doc_i$ compared to all entity pairs generated in NLP domain papers, where the distance between the two entities in the pairs from Doci falls within the top 10% of the distance distribution of entity pairs.

### 3.3.3 Measurement of Novelty Bias Based on Fine-Grained Entity Combinations

Furthermore, from Table 6, we observe an imbalance in the distribution of different types of entities, with Method entities accounting for over fifty percent, followed by Dataset entities at 22%, while Metric entities and Tool entities have lower proportions, accounting for only 5% and 6.6% respectively. To gain a deeper understanding of the novelty bias in papers classified as novel across the categories of methods, datasets, metrics, and tools, an analysis is conducted on the contribution of the types of entity combinations to the novelty of academic papers by measuring the proportion of atypical entity combinations found in novel papers by type. It is known that there are a total of 10 types of entity combinations in papers, including Method-Method, Dataset-Dataset, Tool-Tool, Metric-Metric, Method-Dataset, Method-Tool, Method-Metric, Dataset-Tool, Dataset-Metric, and Tool-Metric combinations.

Formulas (3) and (4) are used to measure the proportion of each of these ten types of entity combinations in novel papers containing atypical combinations, reflecting the bias towards novelty in academic papers across these ten types. For example, if all atypical entity combinations in a novel paper belong to the Method-Method combination type, we can infer that Method novelty contributes the most to the novelty of the paper, indicating a stronger bias towards Method novelty in terms of novelty orientation.

$$\text{Ratio of } E_a\text{-}E_a = \frac{n_{E_a\text{-}E_a}}{m} \quad (3)$$

In Formula (3), $E_a\text{-}E_a \in$（method-method,dataset-dataset,tool-tool,metric-metric）

$$\text{Ratio of } E_a\text{-}E_b = \frac{n_{E_a\text{-}E_b}}{m} \quad (4)$$

In Formula(4), $E_a\text{-}E_b \in$（method-dataset,method-tool,method-metric,dataset-tool,dataset-metric,tool-metric）

**Table 6. Proportion of different types of entities**

| Entity type | Number of entities | Percentage of entities |
|---|---|---|
| Method | 97215 | 64.37% |
| Dataset | 33260 | 22.03% |
| Tool | 12980 | 6.6% |
| Metric | 7563 | 5% |
| Total | 151018 | 100% |

## 4 Result

This study utilized data from three major conferences in the field of Natural Language Processing (NLP) as its primary source. It first analyzed the publication conditions of different types of team institutional compositions at these conferences, including academic institutions, industrial institutions, and mixed academic and industrial institution. The study focused on investigating the differences in novelty among academic papers from various institution types. Through comparative analysis, the study explored the differences in novelty under different types of team institutional compositions. Moreover, by controlling for other potential factors influencing academic paper novelty, regression models were employed to study the relationship between institution types and the novelty of academic papers. The research aimed to examine the impact of institution types on the novelty of academic papers. Additionally, using fine-grained entity combinations, the study investigated the preferences of highly novel papers from different institution types in terms of four aspects of novelty: methods, datasets, tools, and metrics.

### 4.1 Analysis of Publication Conditions of Different Types of Team Institutional Compositions

Based on the aforementioned criteria, the 1901 institutions to which the authors belong were categorized into two main groups: academic institutions and industrial institutions. Further categorization of academic paper institutions led to three groups: academic institutions, industrial institutions, and mixed academic and industrial institutions. As depicted in Figure 2(b), pure academic institutions dominate (69%, 10345), followed by combined academic and industrial institutions (16%, 3176), with pure industrial institutions accounting for 8% (1290). Furthermore, a binary classification based on the first author's institution was conducted to determine the distribution of academic paper institution types, as shown in Figure 2(a), where academic institutions prevail (86.4%, 12803) over industrial institutions (13.6%, 2006).

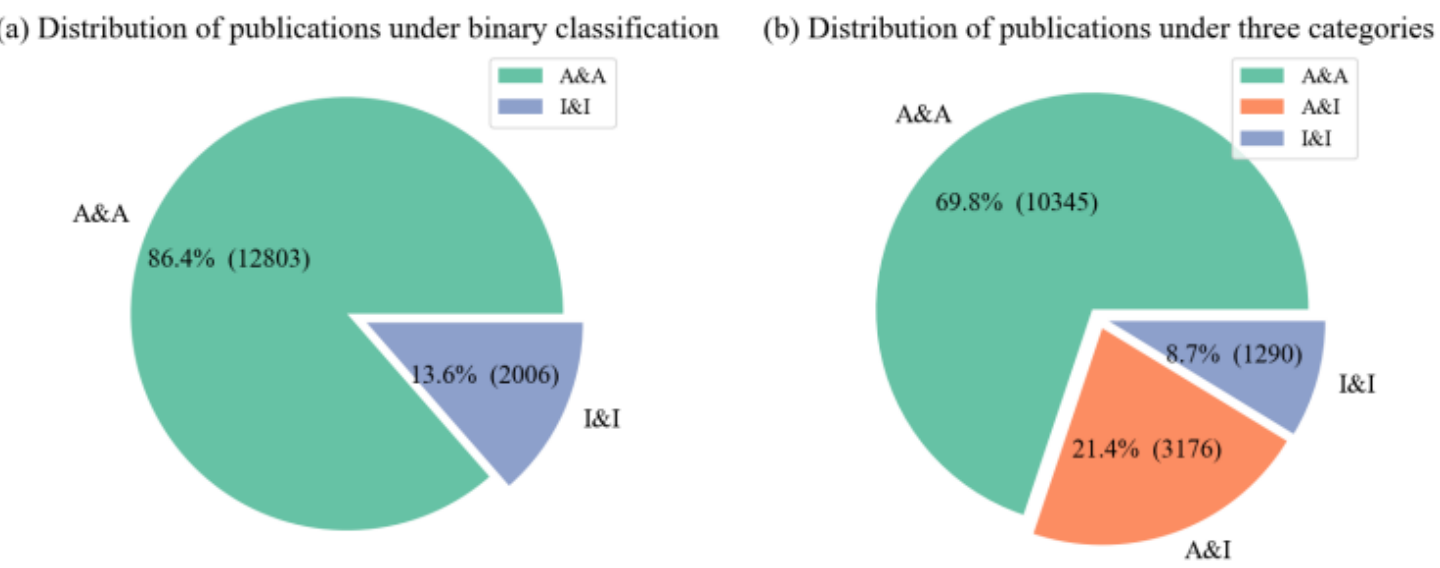

**Fig. 2. Distribution of different institution types**

The analysis of the evolution in publication volume by different types of team institutional compositions over time is presented in Figure 3.

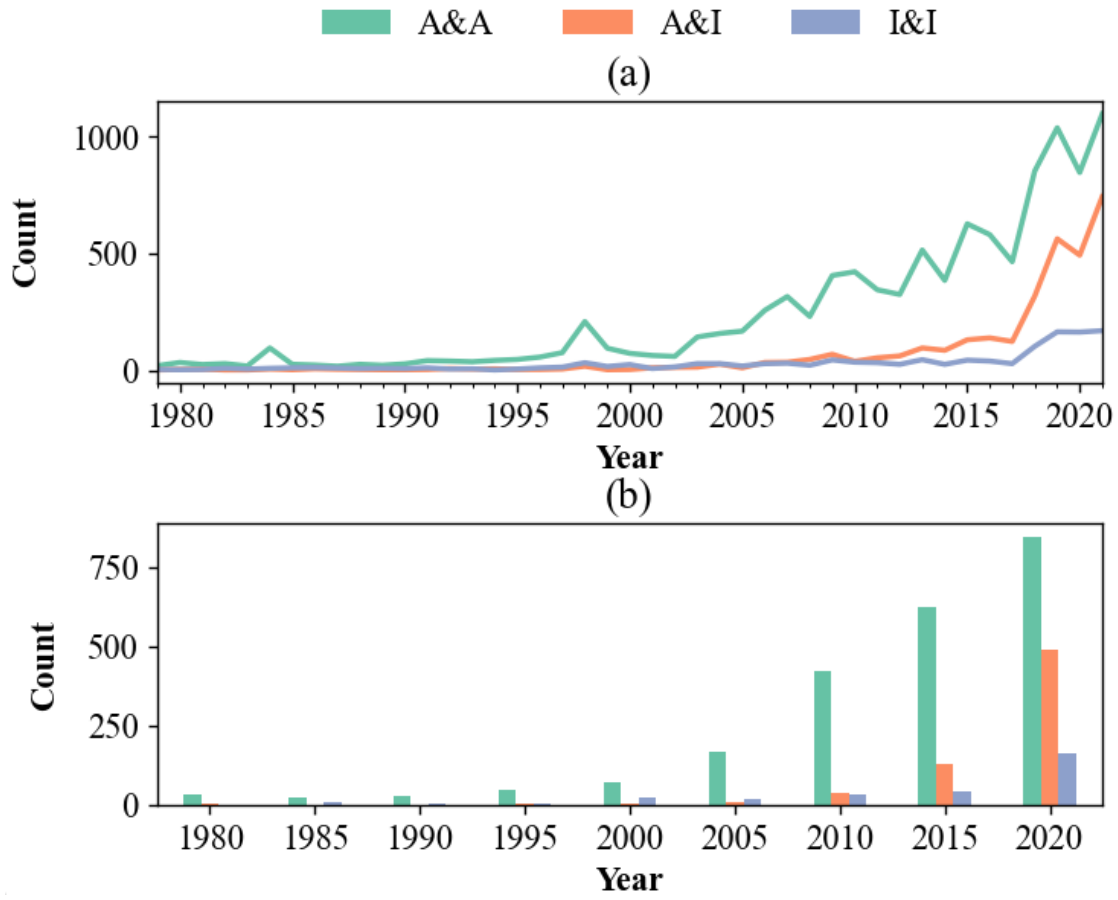

**Fig. 3. Distribution trends of different institution types over time**

The data indicates a general increase in the number of papers published by all types of institutions over the years, particularly with a significant surge in total publications since the early 2000s. This growth may be attributed to rapid technological advancements and increased interest in natural language processing during that period. Additionally, data loss in earlier years and potential omission of some papers during dataset collection may have contributed to the observed rise in publication numbers. Examining the trends in publication volume by different institution types reveals that academic institutions have consistently been the main contributors, occupying the top position in terms of publication volume. However, collaborative papers between industrial and academic institutions have seen a remarkable surge, surpassing the growth rate of academic institution papers, from zero publications in 1979 to 742 publications in 2021. Starting from 2010, there has been a continuous increase in the proportion of academic papers co-authored by industrial institutions each year, with more collaborative papers appearing in top-tier conferences. In alignment with the context provided earlier, many countries are increasingly shifting towards academic-industry collaboration initiatives, advocating for universities and businesses to collaborate to break technological barriers together, which has facilitated the rapid growth in the number of collaborative publications between academic and industrial institutions. Furthermore, while industrial institutions have historically lagged behind in publication numbers, there has been a noticeable increase in the last five years. This growth could be attributed to pure industrial entities placing greater emphasis on practical application results compared to academic institutions' lesser focus on publishing papers. With technological advancements, many industrial institutions are recognizing the importance of research and innovation, indicating a potential shift towards more academic paper publications in the future. To visually represent the trend of industrial institutions' involvement in academic paper publications more intuitively, Figure 4 illustrates the proportion of papers published each year by all organizations attributed to industrial institutions, showing an upward trend from 2018 to 2020. Although industrial institutions still occupy a relatively small portion compared to academic institutions, the upward trend is evident.

In conclusion, the number of papers published in the field of natural language processing continues to grow steadily, with a growing interest in collaboration between academic and industrial sectors likely becoming a major trend in the future. While the quantity of papers published by industrial institutions has been relatively low, there is a clear development trend indicating potential growth opportunities.

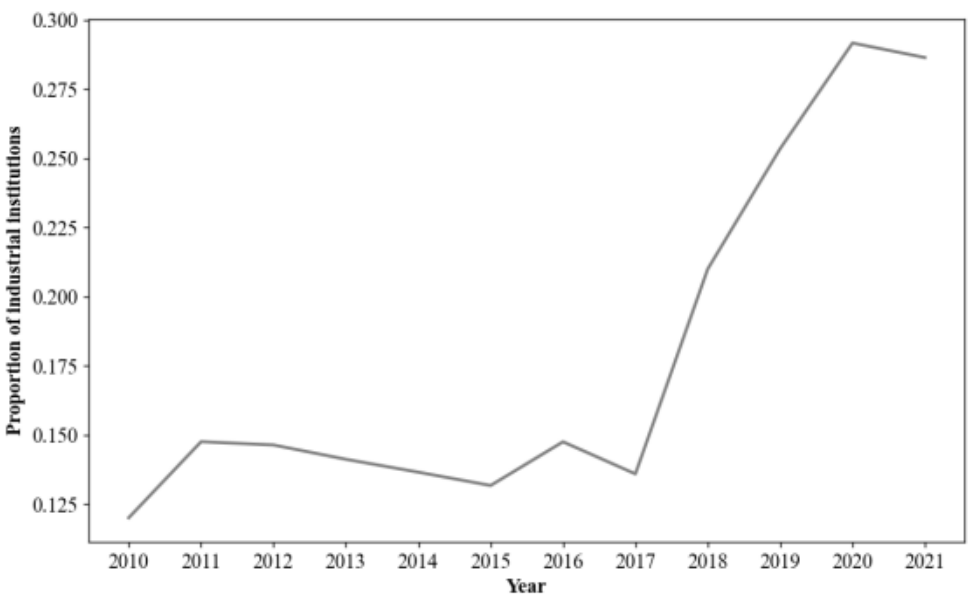

**Fig. 4. Percentage of industry organizations participating each year**

## 4.2 Team Composition and Novelty of Academic Papers

As mentioned above, there are significant differences in the publication output among the three types of institutional compositions, particularly in terms of the number of papers published. The main objective of this paper is to explore the relationship between different types of team institutional compositions and the novelty of academic papers. This study measures the novelty of academic papers based on fine-grained knowledge entity combinations of methods, data, tools, and metrics. Given the substantial differences in the quantities among the four types of entities, after measuring whether academic papers are novel based on all entity combinations, this study further discusses the pairwise combinations of the four types of entities in novel papers. The aim is to delve into the fine-grained novelty bias in novel papers and gain a deeper understanding of the differences in bias towards fine-grained novelty among different types of team institutional compositions.

### 4.2.1 Distribution of Novelty Scores in Academic Papers

Based on the theory of combinative innovation, the novelty of academic papers is measured using extracted method entities. The measurement results are index values ranging from 0 to 1, where values closer to 1 indicate a higher proportion of atypical knowledge combinations, suggesting a likelihood of being highly novel papers, while values closer to 0 indicate a smaller proportion of atypical knowledge combinations, indicating more conservative papers. Figure 5 illustrates the distribution of novelty scores of papers from the three major conferences measured using the aforementioned method, with frequency represented on the Y-axis. From the graph, we observe that the majority of papers have novelty scores concentrated between 0.1 and 0.2, with only a small portion of papers having novelty scores greater than 0.5. This indicates that most papers are still characterized by lower novelty, reflecting conservative research approaches.

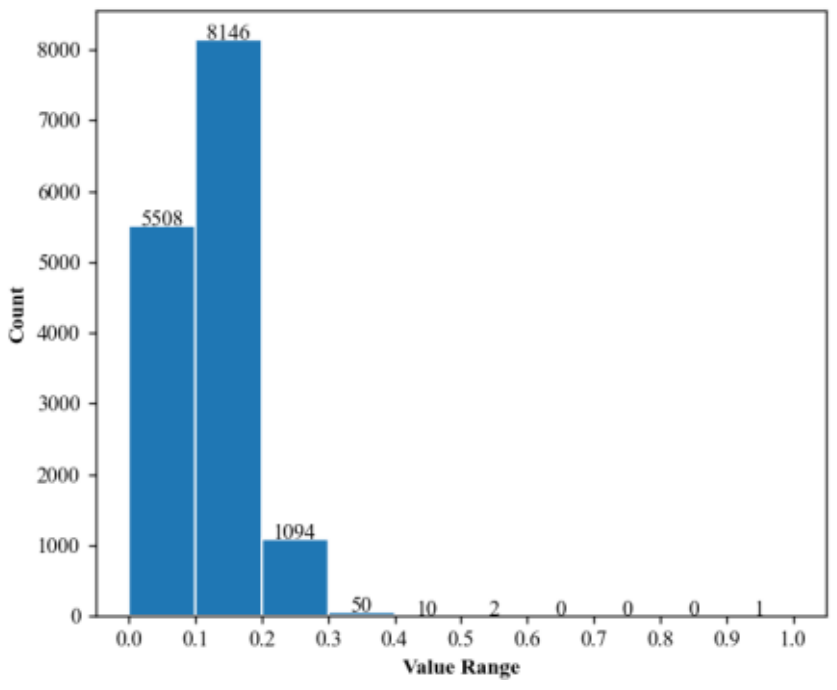

**Fig. 5. The distribution of ACL paper's novelty scores**

Subsequently, we analyzed the trend in novelty over time. As shown in Figure 6, there was a notable and consistent increase in novelty scores between 1995 and 2000. This upward trend can be attributed to several factors. Firstly, advancements in technology, such as the introduction of new algorithms and models, have driven rapid progress in the field. Secondly, enhancements in computational capacities have allowed researchers to work with larger and more complex datasets, facilitating more comprehensive studies. Interdisciplinary collaborations have brought in fresh perspectives and methodologies, while increased international cooperation has fostered cutting-edge research on a global scale. The growing demand for practical applications in natural language processing has motivated researchers to propose innovative solutions. Collectively, these factors have stimulated innovation and advancement in NLP, establishing a robust foundation for future research endeavors.

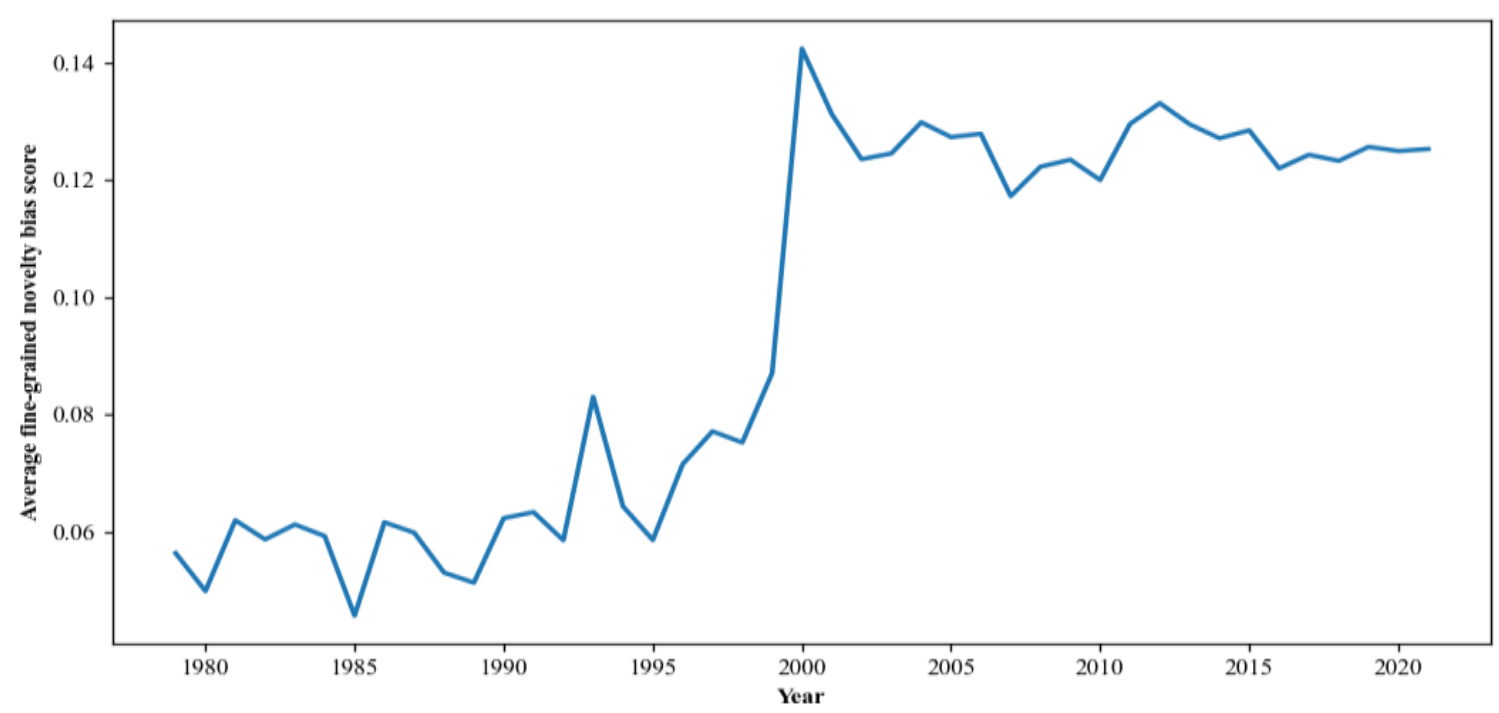

**Fig. 6. Trends in novelty scores of academic papers over time**

### 4.2.2 Different Institutional Compositions and Novelty of Academic Papers

### (1) Introduction to variables

According to Figure 7, significant differences in the novelty of academic papers are observed among different types of team institutional compositions. It is found that academic institutions and mixed academic and industrial institutions produce papers with higher novelty compared to papers from industrial institutions. To further investigate the relationship between different institution compositions and the novelty of academic papers, this study employs regression analysis. To ensure the reliability of the results, controls are first implemented for year and task variations. The novelty of academic papers is measured in two ways: firstly, by categorizing papers as highlynovel if their novelty score falls within the top ten percent (assigned a value of 1) and non-highly novel otherwise (assigned a value of 0); secondly, by utilizing the continuous variable of the novelty score as the dependent variable. Different institution compositions are treated as independent variables, with the industrial institution as the reference group to analyze the relationship between types of team institutional compositions and the novelty of academic papers. Furthermore, to enhance the robustness of the results, a second regression model is constructed by introducing the balance ratio of academic-industrial institution collaborations as an independent variable. This study considers papers as the basic research unit, where factors such as the number of institutions in a team, the influence of the institutions themselves, and the presence of international collaborations all impact the novelty of papers. Therefore, these variables are controlled for in the regression analysis to more accurately estimate the core variables. The variables included in the regression analysis are presented in Table 7.

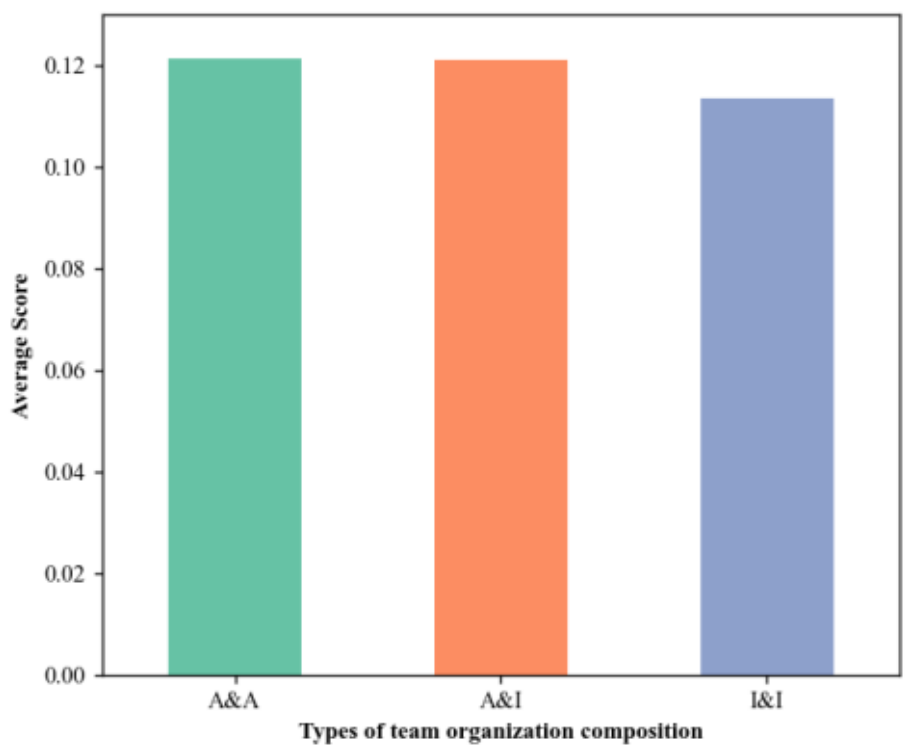

**Fig. 7.Novelty of papers of different institution types**

**Table 7. Regression analysis of variables and their interpretations**

| Variable | Explain |
|---|---|
| Paper novelty1 | To determine whether a paper is a high novelty paper, the variable is set to 1 if it is a high novelty paper, otherwise it is 0. |
| Paper novelty2 | The novelty score of an academic paper indicates the level of novelty, where a higher score indicates a higher level of novelty and a lower score indicates a lower level of novelty. |
| The type of institution | The different types of institutions in the paper are academia(A&A), industry(I&I), and a mix of academia and industry(A&I) |
| Top-level organization or | Set it to 1 when the paper contains high-level institutions, and 0 |
| not | otherwise. (Academic institutions are considered to be high-level institutions in computer science rankings, and industrial institutions are considered high-level institutions in the world's top 500 list) |
| International cooperation or not | Set it to 1 when there is cross-border collaboration in the paper, and 0 when there is no other paper |
| Team size | The number of people on the team |
| Task | The type of task to which the paper belongs |
| Year | Year of publication |

### (2) Regression Analysis

In Model (1) of Table 8, the dependent variable is whether the paper is highly novel, a binary variable. It is found that relative to I&I, A&I has a significantly higher positive impact on the novelty of academic papers (coefficient = 0.062, $P < 0.01$), and A&A also has a positive impact on the novelty of academic papers (coefficient = 0.05, $P < 0.01$). In Model (2), a set of control variables is introduced. A&I and A&A still have positive impacts compared to I&I, and they remain significant at the 1% level. To control for any time-varying changes at the year level, we added year fixed effects in the third column. A&I and A&A still have statistically significant positive impacts compared to I&I at the 1% level. Additionally, when combining task fixed effects and year fixed effects, A&I and A&A still have statistically significant positive impacts compared to I&I at the 1% level. This suggests that compared to I&I, A&I and A&A have positive impacts on the novelty of academic papers.

In Model (5) of Table 9, it is found that relative to I&I, both A&I and A&A have a significantly higher positive impact on the novelty of academic papers (coefficient = 0.062, $P < 0.01$; coefficient = 0.05, $P < 0.01$). In Model (6), additional variables are added, and in Model (7), year fixed effects are controlled. Model (8) further controls for task fixed effects. The results consistently show that compared to I&I, both A&I and A&A collaborations have positive impacts on the novelty of academic papers.

From the results of these two regression analyses, it is found that relative to the industrial sector, mixed institutions involving both academic and industrial sectors have the highest positive impact on the novelty of academic papers, with academic institutions also showing significant positive impacts. The regression coefficients of the control variables suggest a significant positive correlation between the inclusion of high-level institutions and international cooperation in teams and the novelty of academic papers. These conclusions are consistent with many previous studies (Lee et al., 2015; Xu et al., 2022).

**Table 8. Regression estimation of the relationship between team institutional composition and novelty score 1**

| Paper's Novelty1 | Model(1) | Model(2) | Model(3) | Model(4) |
|---|---|---|---|---|
| The type of institution | | | | |
| A&A | 0.05*** | 0.05*** | 0.043*** | 0.0389 *** |
| | (0.007) | (0.0071) | (0.0071) | (0.0071) |
| A&I | 0.062*** | 0.0445*** | 0.04*** | 0.0437*** |
| | (0.0084) | (0.0091) | (0.01) | (0.009) |
| Top-level organization or not | | 0.0132** | 0.0135** | 0.0125** |
| | | (0.006) | (0.0061) | (0.0062) |
| International cooperation or not | | 0.0116* | 0.0114* | 0.0118** |
| | | (0.007) | (0.007) | (0.007) |
| Team size | | 0.0083*** | 0.007*** | 0.0087*** |
| | | (0.0019) | (0.002) | (0.002) |
| Constant | 0.0516*** | 0.134 | 0.0202 | 0.028 |
| | (0.0063) | (0.01) | (0.0675) | (0.09) |
| Task fixed effect | NO | NO | YES | YES |
| Year fixed effect | NO | NO | NO | YES |
| $R^2$ | 0.0085 | 0.006 | 0.018 | 0.035 |
| Observations | 14812 | 14812 | 14812 | 14812 |

**Note:** In parentheses is a robustness criterion.. * indicates P<0.1, ** indicates P<0.05, and *** indicates P<0.01

**Table 9. Regression estimation of the relationship between team institutional composition and novelty score 2**

| Paper's Novelty2 | Model(5) | Model(6) | Model(7) | Model(8) |
|---|---|---|---|---|
| The type of institution | | | | |
| A&A | 0.0113*** | 0.0114*** | 0.0098*** | 0.0065*** |
| | (0.0015) | (0.0016) | (0.0016) | (0.0015) |
| A&I | 0.0207*** | 0.0139*** | 0.0122*** | 0.01*** |
| | (0.0019) | (0.002) | (0.002) | (0.0019) |
| Top-level organization or not | | 0.0055*** | 0.005*** | 0.0034*** |
| | | (0.0012) | (0.0012) | (0.0012) |
| International cooperation or not | | 0.0041*** | 0.0042*** | 0.0027** |
| | | (0.0014) | (0.0013) | (0.0013) |
| Team size | | 0.0033*** | 0.0026*** | 0.0011*** |
| | | (0.002) | (0.0003) | (0.0003) |
| Constant | 0.101*** | 0.094*** | 0.104*** | 0.047** |
| | (0.0015) | (0.002) | (0.011) | (0.016) |
| Task fixed effect | NO | NO | YES | YES |
| Year fixed effect | NO | NO | NO | YES |
| $R^2$ | 0.0085 | 0.022 | 0.054 | 0.119 |
| Observations | 14812 | 14812 | 14812 | 14812 |

**Note:** In parentheses is a robustness criterion.. * indicates P<0.1, ** indicates P<0.05, and *** indicates P<0.01

### 4.2.3 Differences in Novelty of Academic Papers Based on Fine-Grained Entity Composition in Different Institutional Compositions

As discussed earlier, there are significant differences in the novelty of papers among different types of team institutional compositions, mixed academic and industrial institutions and academic institutions demonstrating higher novelty. This study measures the novelty of academic papers based on fine-grained entity compositions. After determining the novelty of academic papers, the proportion of atypical entity combinations in novel papers is further analyzed to explore the bias towards novelty. This analysis aims to provide a deeper understanding of the differences in novelty bias among different types of team institutional compositions.

**(1) Contribution of Fine-Grained Entity Combinations to Novelty**

After obtaining the distribution of overall novelty scores of academic papers, a more detailed measurement of novelty in academic papers is conducted based on the methodology introduced in the research. The proportions of ten types of entity combinations in novel papers are measured to assess their contributions to unique entity combinations in these papers. Figure 8 illustrates the proportion of each type of entity combination. As shown in Figure 8, consistent with the distribution pattern of entities, method-related entity combinations have a higher representation among the entity combinations analyzed. Among these, method-indicator entity combinations make the most significant contribution, followed by method-method and method-data entity combinations, reflecting the emphasis on technical applications in natural language processing. In contrast, entity combinations related to indicators exhibit relatively lower contributions to the novelty of novel papers. This is because the indicators used in research are often standardized and play an evaluative role in assessing model efficiency, with less room for innovation specifically targeting indicators. Moreover, tools in the field of natural language processing primarily focus on dataset handling, analysis, and framework construction for model methods, leading to a lesser impact on the novelty of novel papers. In the field of natural language processing, data entities are often combined with method entities to demonstrate novelty, such as processing old data with new methods or vice versa.

Upon reviewing relevant papers, it was found that Liu et al. (2019) introduced a sentence encoder capable of continuously learning features from new corpora while maintaining performance on past corpora. The key innovation lies in proposing new sentence encoder methods tailored to different corpora. Our analysis indicates that the novelty of this paper is predominantly driven by atypical method-data entity combinations, followed by method-method entity combinations. In another study, Elliott and Keller (2013) focused on incorporating visual dependency representations to capture object relationships in images and utilized Amazon Mechanical Turk for annotating image data. Our findings reveal that atypical method-data entity combinations make the highest contribution to novelty, followed by data-tool entity combinations. These instances highlight the explanatory power of the contribution measurement metrics proposed in this research regarding novelty.

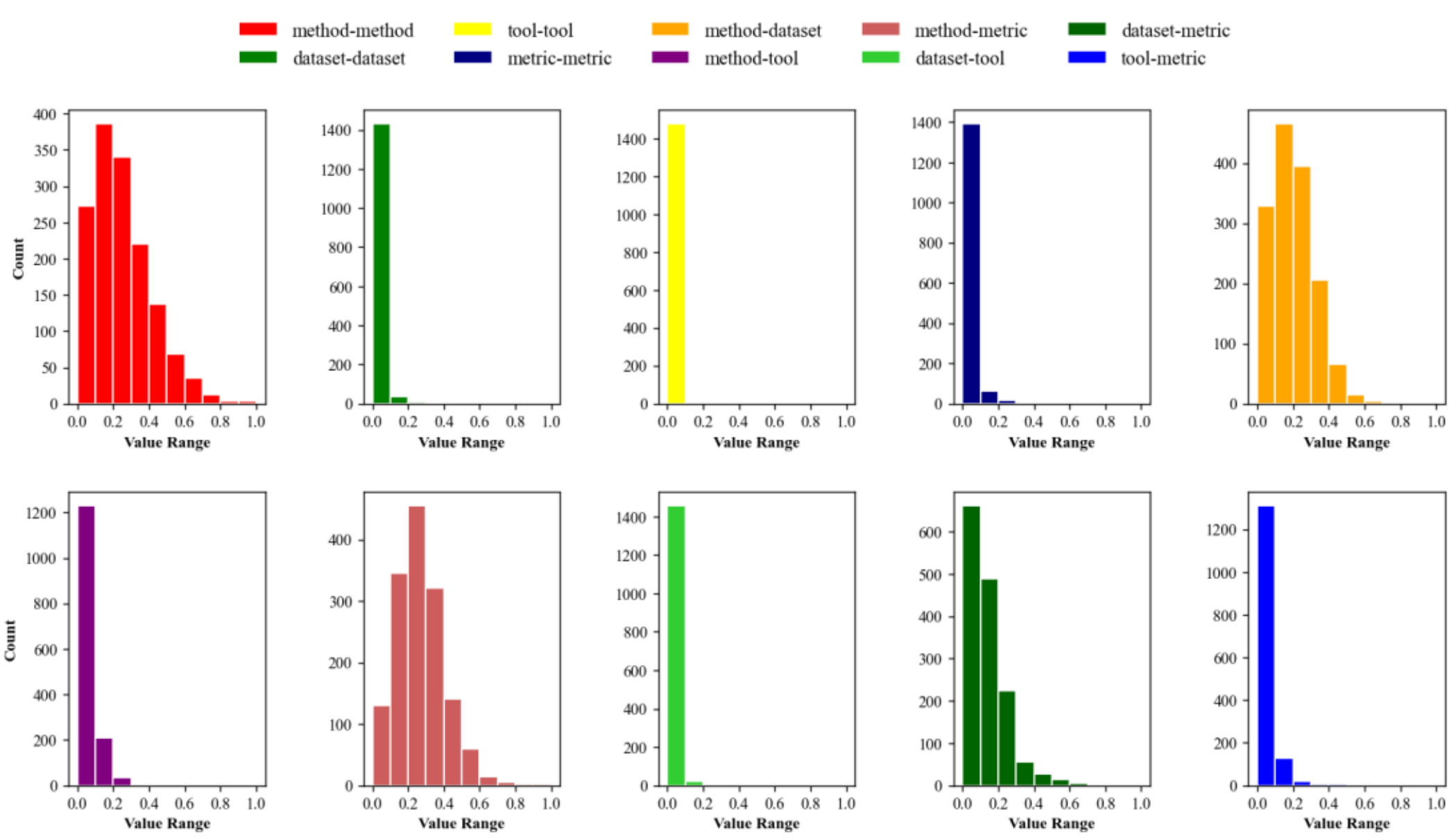

**Fig. 8. Fine-grained novelty biases the score distribution**

### (2) Analysis of Differences in Fine-Grained Novelty Contributions among Different Institutional Compositions

Building upon the findings from the preceding section, this section delves into the nuanced differences in novelty contributions across various team institutional compositions. By referencing Figure 9, the study scrutinizes the proportions of ten types of entity combinations that contribute to the novelty in papers published by different institutional compositions. It is observed that entities associated with methods exert the most significant influence on novelty, suggesting a prevalent focus on method-related innovations irrespective of the institutional composition in the field of natural language processing. Overall, method-related entities constitute a substantial portion of the novelty contribution to novel papers, accounting for 76.61% of the total.

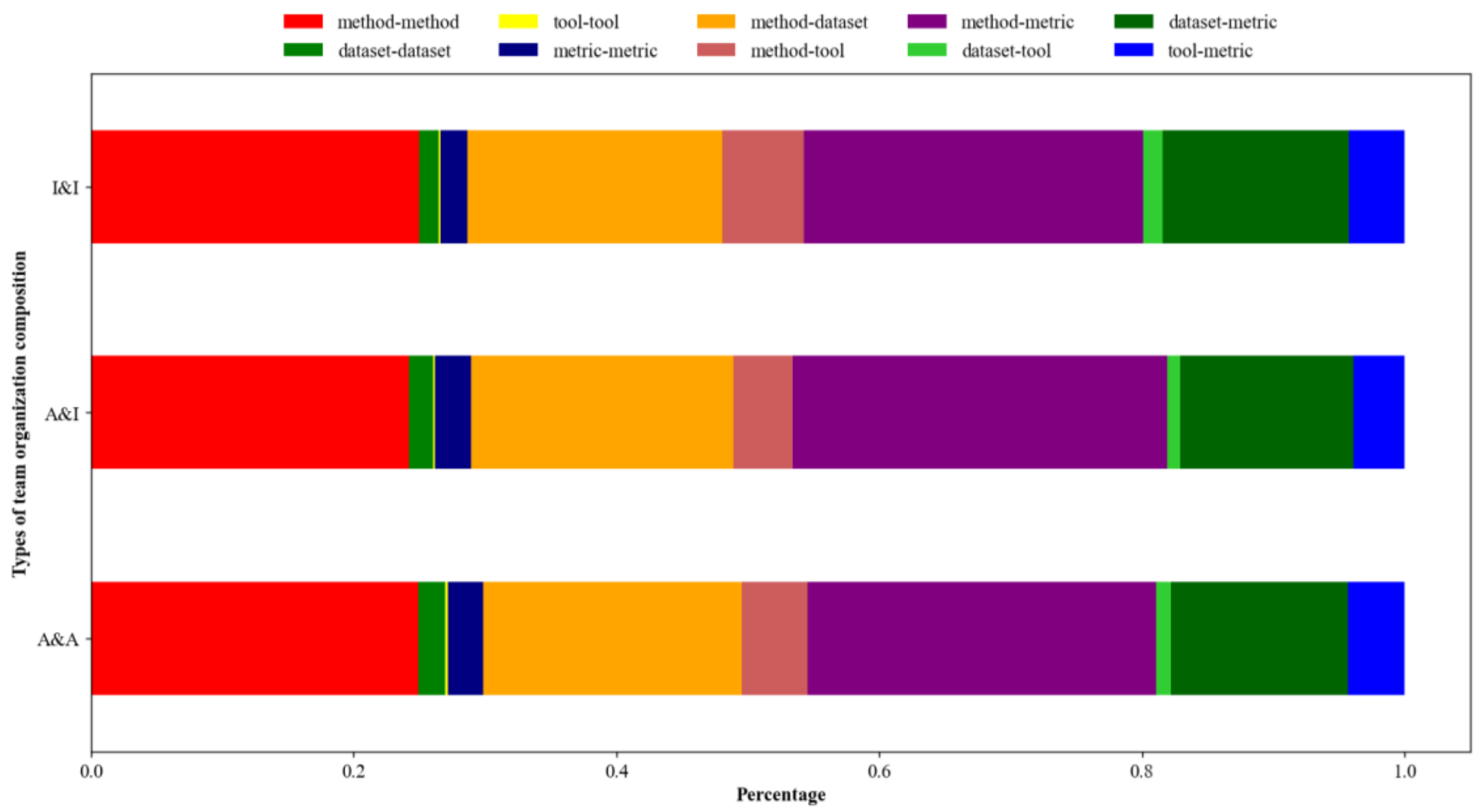

**Fig. 9. Proportion of the contribution of the combination of entities of different institution types to novelty**

To explore the variations in fine-grained novelty contributions among different team institutional compositions, this study further employed one-way analysis of variance for analysis. The results, depicted in Figure 10, unveil significant disparities between distinct institutional compositions concerning novelty contributions in method-tool and method-indicator combinations ($P < 0.1$, $P < 0.05$). Specifically, mixed academic and industrial institutions demonstrate notably higher novelty contributions in the method-indicator combination compared to other institutions, whereas industrial institutions exhibit significantly higher novelty contributions in the method-tool combination. This phenomenon can be attributed to the enhanced feasibility of interdisciplinary collaboration within mixed academic and industrial institutions, enabling the fusion of methods and indicators from diverse domains. Such interdisciplinary cooperation may introduce fresh perspectives and research methodologies, leading to a preference for method-indicator combinations in terms of novelty. Conversely, industrial institutions are directly confronted with the practical demands of real-world applications, thereby displaying a propensity to leverage innovative tools for tackling tangible problems. This pragmatic orientation towards application might propel industrial institutions towards innovation in the method-tool combination.

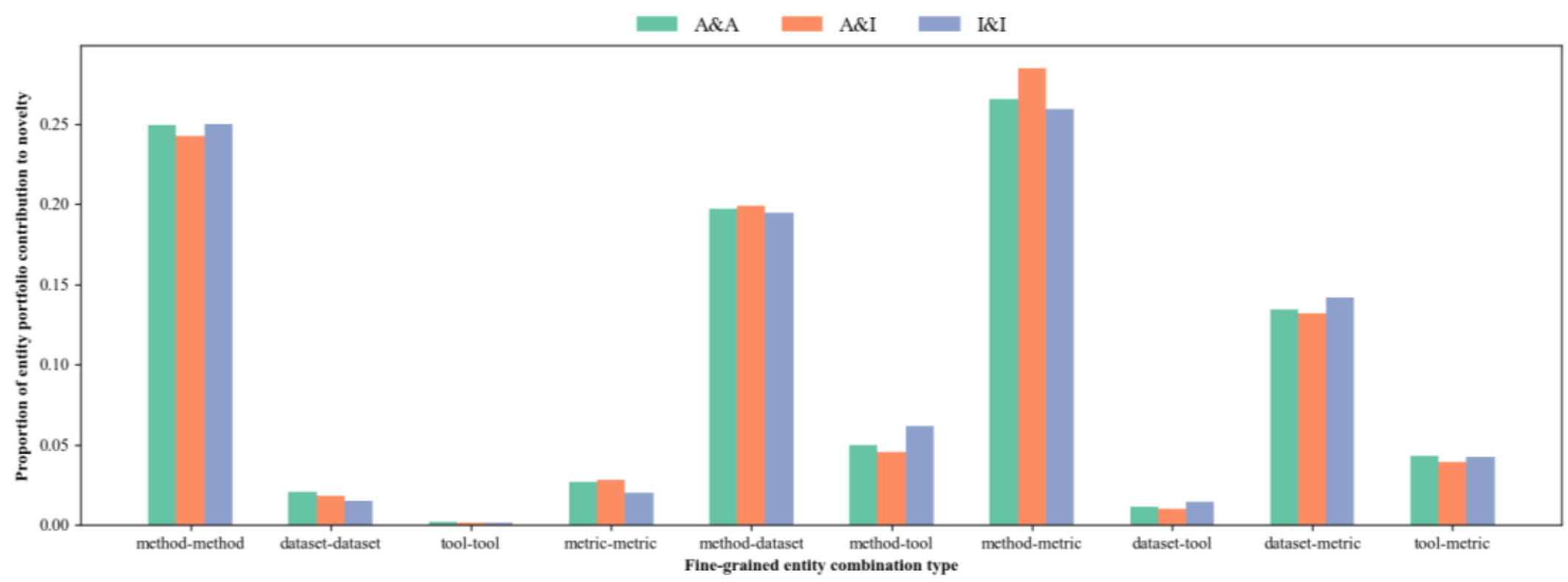

**Fig. 10. The contribution of fine-grained entity combinations of different institutional types to novelty**

## 5 Case Study

Taking the paper "Knowledge Supports Visual Language Grounding: A Case Study on Colour Terms" by Simeon et al. (2020) as an example. This paper mainly explores the supportive role of world knowledge in human cognition on object color perception. The summary of the innovation points of this paper in the original text as indicated by the underlined sentence in Figure 11.

helps to perceive their colour in certain contexts. We go beyond previous studies on colour terms using isolated colour swatches and study visual grounding of colour terms in realistic objects. Our models integrate processing of visual information and object-specific knowledge via hard-coded (late) or learned (early) fusion. We find that both models consistently outperform a bottom-up baseline that predicts colour terms solely from visual inputs, but show interesting differences when predicting atypical colours of so-called colour diagnostic objects. Our models also achieve promising results when tested on new object categories not seen during training.

**Fig. 11. Innovation Description in Case Study**

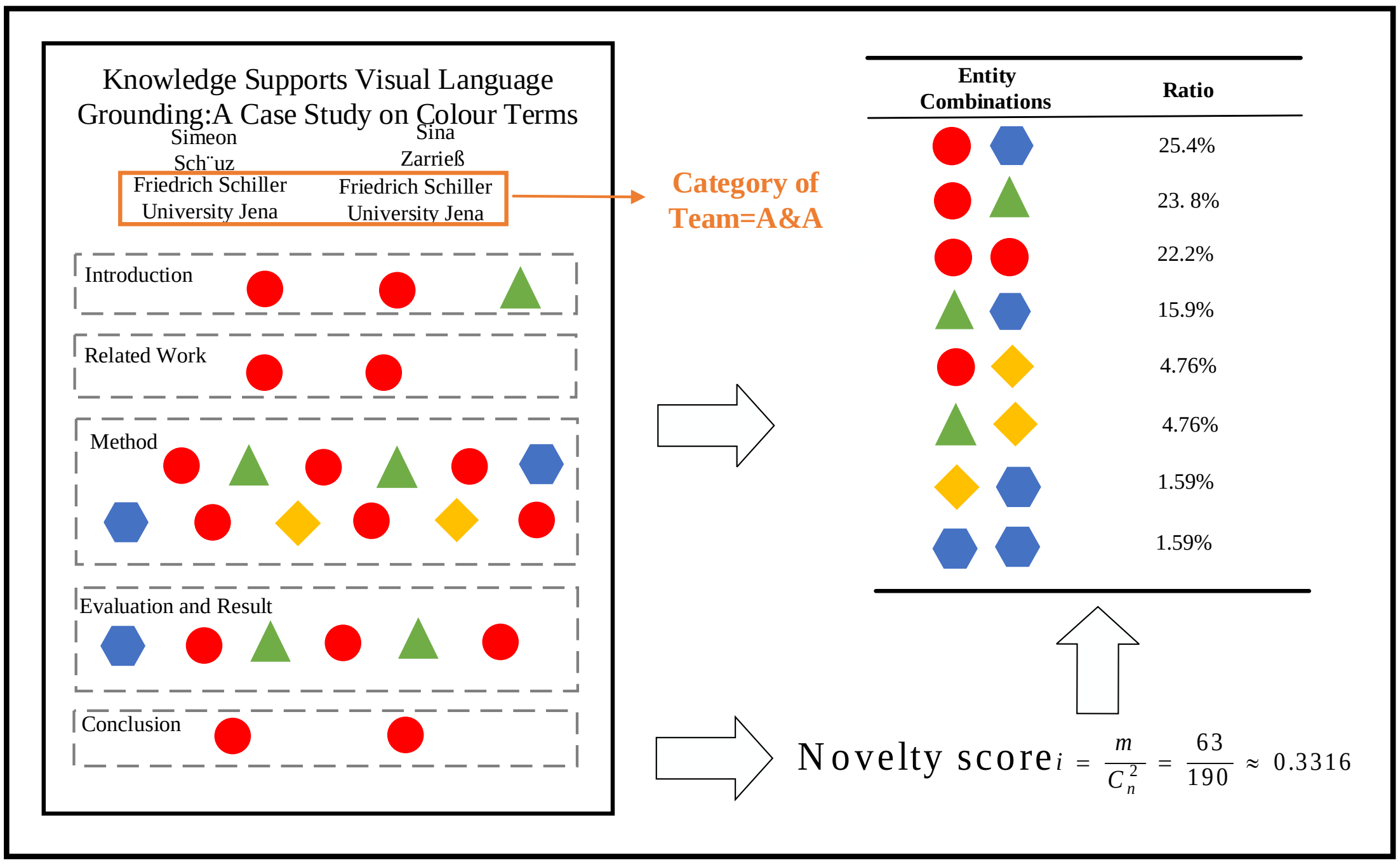

**Fig. 12. Fine-Grained Novelty Bias Result Analysis Example Diagram**

As illustrated in Figure 12, this study examines the novelty of the paper mentioned. Initially, without considering fine-grained entity types, the novelty score of the paper is calculated to be 0.3316 using a method that measures novelty through knowledge entity combinations. This indicates that the novelty score of this paper ranks in the top 10% of the dataset, classifying it as a novel paper. Upon introducing the perspective of fine-grained knowledge entity combinations and analyzing atypical entity combinations in the paper, it is revealed that the novelty contribution consists of 22.2% from method-method entity combinations, 1.59% from indicator-indicator combinations, 23.8% from method-data combinations, 25.4% from method-indicator combinations, 4.76% from method-tool combinations, 15.9% from data-indicator combinations, 4.76% from data-tool combinations, and 1.59% from indicator-tool combinations. Compared to measuring novelty without considering fine-grained knowledge entities, employing fine-grained knowledge entities further enhances our understanding that the innovation of the paper primarily lies in method-related combinations, even without delving into the literature. Based on the percentages of different entity combinations, it can be inferred that the content possibly involves improving methods on existing datasets and evaluating new methods using multiple indicators. Furthermore, connecting the fine-grained novelty to the institutional composition of the paper's team, where both authors originate from academic institutions, indicates an academic team structure. The novelty predominantly stems from method-data entity combinations, along with method-method and method-indicator entity combinations, aligning with the findings from Figure 10 regarding fine-grained novelty in academic institutions.

By analyzing this paper, we can discover that measuring the novelty of academic papers using fine-grained entity combinations helps us understand more than just the degree of novelty. It allows us to identify which types of entity combinations contribute most significantly to the novelty of the paper, giving us a rough idea of the types of novelty present in the paper. Furthermore, by correlating this analysis with the composition of academic paper teams, we can uncover differences in novelty preferences among institutions of various types.

# 6 Discussion

In this section, we discuss the significance and limitations of our study. In Section 6.1, we analyze the value of this research from both theoretical and practical perspectives. In Section 6.2, we address existing constraints and propose corresponding improvement suggestions.

## 6.1 implication

The paper discusses the significance of our research from theoretical and practical standpoints.

**(1) Theoretical Implications:**

This study presents a novel perspective and methodology for investigating the relationship between team composition and the novelty of academic papers. Unlike previous research, this study focuses on the composition of different institutions in academia and industry, revealing differences in their impact on paper novelty. By analyzing novelty across various types of institutions and the contributions of fine-grained entity combinations, the study refines novelty measurement from traditional binary or scalar approaches to diverse novelty biases. This sophisticated approach to measuring novelty offers researchers in academia and industry a comprehensive evaluation tool, enabling a nuanced understanding of the unique features of papers and providing valuable insights for future research. The expansion of this research methodology contributes to the advancement of novelty research, fostering enhanced collaboration and communication between academia and industry in the realm of innovation.

**(2) Practical Implications:**

This study delves into the relationship between institutional composition and the novelty of academic papers in the field of natural language processing. The findings reveal a discernible relationship between institution type and academic paper novelty, for industrial institutions, the probability of publishing novel papers in collaboration with academic institutions was higher, underscoring the importance of diverse institutional collaborations for knowledge generation. Further analysis demonstrates that mixed academic and industrial institutions significantly outperform other institutions in terms of method-indicator combination novelty contributions, while industrial institutions exhibit heightened novelty contributions in method-tool combinations relative to other types. These results underscore the distinctiveness of novelty biases among different institutional compositions, emphasizing the critical role of interdisciplinary collaboration and advocating for increased cooperation and communication between academia and industry to drive innovative knowledge creation and practical applications.

### 6.2 limitation

Indeed, this study is subject to certain limitations. These include the restricted data sources, constraints in measuring the novelty of academic papers, and associated analyses. Firstly, regarding data limitations, the study solely relied on data from the three prominent conferences in the field of natural language processing. Consequently, this dataset may not be fully representative of all research endeavors within the natural language processing domain or capture the comprehensive spectrum of relationships therein. Moreover, the classification of author affiliations into academic institutions, academic-industry hybrid institutions, and industrial institutions resulted in a substantial variance in the number of papers across these categories, with a predominant presence of academic papers and a notably underrepresented proportion of industrial papers. Such data distribution imbalances could potentially introduce inaccuracies into the analysis results. Secondly, although the measurement of academic paper novelty involved the combination of various fine-grained entity types, the study did not constrain these entity combinations to specific contexts or distinctly delineate the interpretations of entity combinations across different structural sections. By pairing entities throughout the entire document without specific contextual distinctions, there exists a possibility that the reliability of measuring the novelty of academic papers may be compromised. Lastly, the study did not delve into the causal relationship between institution type and the novelty of academic papers, presenting a latent limitation in this investigation.

## 7 Conclusion and future works

In this study, we conducted an in-depth exploration of the relationship between institutional composition and academic paper novelty within a specific field. Initially, we categorized scholar team institutions into three types based on authors' affiliations (A&A, A&I, I&I) and employed a sophisticated entity recognition model to identify fine-grained knowledge entities related to methods in NLP papers. Subsequently, we utilized the theory of combinatorial innovation to measure the novelty of academic papers and differentiated papers into novel and non-novel categories based on the novelty measurement results. Furthermore, we conducted a nuanced assessment by combining four types of entities within novel papers to investigate variations in fine-grained novelty bias among papers published by different institutional compositions.

For future research endeavors, we aim to expand the dataset by potentially incorporating patent data sources to enhance the precision of our analyses. Additionally, we seek to optimize our novelty measurement metrics by firstly enhancing the accuracy of scientific entity recognition and secondly integrating structural elements such as document sections into the assessment of academic paper novelty to capture more refined measures of novelty. Finally, we plan to continue investigating the potential causal relationship between institution type and academic paper novelty while intending to validate the outcomes of this study using diverse datasets from other domains. These extensions will contribute towards a comprehensive understanding of the influence of institution type on academic paper novelty and further propel advancements in related research fields.

## Acknowledgements

This study is supported by the National Natural Science Foundation of China (Grant No. 72074113) and the Open Funding Project of Laboratory of ISTIC-Springer Nature Joint Laboratory for Open Science (Grant No. ISN23012).